# A data science and machine learning approach to continuous analysis of Shakespeare's plays

**Charles Swisher, Lior Shamir**

Kansas State University, Unite States of America

*Corresponding author: Lior Shamir    lshamir@mtu.edu

**Abstract**
The availability of quantitative text analysis methods has provided new ways of analyzing literature in a manner that was not available in the pre-information era. Here we apply comprehensive machine learning analysis to the work of William Shakespeare. The analysis shows clear changes in the style of writing over time, with the most significant changes in the sentence length, frequency of adjectives and adverbs, and the sentiments expressed in the text. Applying machine learning to make a stylometric prediction of the year of the play shows a Pearson correlation of 0.71 between the actual and predicted year, indicating that Shakespeare's writing style as reflected by the quantitative measurements changed over time. Additionally, it shows that the stylometrics of some of the plays is more similar to plays written either before or after the year they were written. For instance, Romeo and Juliet is dated 1596, but is more similar in stylometrics to plays written by Shakespeare after 1600. The source code for the analysis is available for free download.

**keywords**
Shakespeare; text analytics; machine learning

## INTRODUCTION
Being one of the most in influential authors in history, the analysis of the stylometrics of William Shakespeare has been a topic of substantial interest. In addition to "traditional" manual analysis, the work of Shakespeare was also analyzed by using mathematical and quantitative approaches [Brainerd, 1980; Nadel and Matsuba, 1990; Bauer and Zirker, 2018]. One of the earliest attempts to apply mathematical analysis to Shakespeare's style was done by Fucks [1952], who studied the frequency of text elements of the writing to further understand the use of language by an author. Williams [1975] analyzed the word length distribution in Shakespeare's plays, and showed that the distribution of words with different lengths showed substantial difference from the work of Bacon. Analysis of Shakespeare plays with the Regressive Imagery Dictionary showed that incongruous juxtapositions is sensitive to time [Derks, 1994]. Lowe and Matthews [1995] used radial a basis function network to show differences in the style of Shakespeare and Fletcher. Another form of quantitative analysis of Shakespeare's work was based on using eye trackers to analyze eye movements of people as they read Shakespeare's sonnets [Xue et al., 2019]. Other work related to the analytics of Shakespeare plays include the visualization of the text [Wilhelm et al., 2013].

In the context of applying computers to analyze Shakespeare's plays, substantial efforts have been made to verify the authenticity of plays attributed to Shakespeare [Merriam, 2009; Rizvi, 2019; Barber, 2020]. Boyd and Pennebaker [2015] used quantitative analysis techniques [Boyd, 2017] to determine whether "Double Falsehood", published after Shakespeare's death, was likely written by Shakespeare. The analysis attempted to identify psychological signatures of three authors by examining the distribution of words they use, their grammar, and the meaning of the words, and then compared those signatures to the text of the unknown play. Elliott and Valenza [2010a,b] applied stylometric analysis to determine whether and what parts of "Sir Thomas More" and "Edward III" were written by Shakespeare, and showed that while some parts are much more likely to be written by Shakespeare, the probability that the entire plays were written by Shakespeare is low. Stylometric analysis was also used to identify gender diferences is Shakespeare's characters [Savoy, 2022].

Some previous work was focused on the changes in Shakespeare's style in the temporal domain, using quantitative analysis aiming at the pro ling of stylistic change over time [Brainerd, 1980; Taylor, 1987; Moscato et al., 2022], and identifying dates of compositions by analysis of the style [Forsyth, 1999; Stamou, 2007]. For instance, regression analysis of Shakespeare's style was used to estimate the dates of creation of the plays [Moscato et al., 2022]. Multivariate analysis of Shakespeare plays showed quantitatively that the writing style is sensitive to the date of composition, and some exceptions can be explained by multiple authorship [Brainerd, 1980]. Authorship of the plays of Shakespeare, who often collaborated with other authors, is also a question that was addressed through the use of quantitative and computer analysis [Craig and Kinney, 2009].

While substantial work has been done on quantitative analysis of Shakespeare plays, less work has been done by using machine learning. Here we applied a comprehensive quantitative machine learning analysis of Shakespeare's plays to identify elements in Shakespeare's writing that changed over time. The analysis also allows to profile the plays by comparing the actual estimated year of the writing with the year of which the content of the play fits, as predicted by the machine learning system.

Unlike previous analyses of Shakespeare, here we apply a data science approach that is not hypothesis-driven. That is, the text of Shakespeare's plays is analyzed without making any prior assumptions, or testing specific questions. For that purpose, machine learning is used to examine and identify complex patterns of style that cover a large number of text and writing elements. The combination of elements measured in the different plays allows to identify similarities between plays and continuous changes without necessarily making prior assumptions on the specific style elements that change over time.

The rest of the paper is organized as follows: Section I describes the data that was used in the analysis, Section II describes the data analysis and machine learning methods, and Section III discusses the results, including the similarities and differences between different plays and specific elements that change across plays or periods of time. The last section briely discusses the conclusions regarding the analysis of Shakespeare work, the limitations of the analysis, future work, and the potential impact of data science on the digital humanities in general.

**I DATA**

Shakespeare is largely believed to have written 38 plays [Zesmer, 1976]. The dataset used for the experiment consists of all 38 plays attributed to Shakespeare. Data of each play is a text file of the plays, collected in plain text format from "The Complete Works of William Shakespeare", available at http://shakespeare.mit.edu. "Edward III", which is believed to be partially written by Shakespeare [Elliott and Valenza, 2010a,b], was not included in the dataset.

To avoid analyzing pieces of text that are not part of Shakespeare's work, all headers and footers were removed from the files, including any preface material and the name of the play.

For each play, the year in which the play was written was also collected. As the exact dates are unknown, a single year was selected from the range of years the play was most likely written, as was analyzed and provided by the Royal Shakespeare Company[1].

## II METHOD

Each of the text files of the plays was processed using the Unified Data Analysis Tool (Udat) [Shamir, 2020], that works with the Core Natural Language Processing (CoreNLP) library [Manning et al., 2014]. Udat is a comprehensive text analysis tool that extracts 298 numerical text descriptors from each text file. Unlike some document classiers, Udat is not based on the detection of certain words that happen to be more frequent in the text, but on the stylistic elements as re ected by a combination of numerous measurements from the text [Shamir, 2020; Rosebaugh and Shamir, 2022].

The text measurements include basic statistics such as the average, standard deviation, and histograms of the words length and sentence length. Other basic statistics is focused on the frequency of punctuation characters. The analysis also measures the diversity of words, the homogeneity of the appearance of words throughout the text, and the frequency and length of quotations in the text as described in detail in [Shamir, 2020].

By using the CoreNLP library [Manning et al., 2014], the distribution of parts of speech is also analyzed. That allows to measure the frequency of different parts of speech such as nouns, verbs, pronouns, etc. The Discrete Fourier Transform (DFT) is applied to measure repetitive patterns in the use of different parts of speech. Another aspect that is measured from the text is the sentiments expressed in it, including the variations of the sentiments throughout the text as explained in [Shamir, 2020].

The automatic readability index [Smith and Senter, 1967] and the Coleman-Liau index [Coleman and Liau, 1975] measure the reading level of the text as defined by these indices. The distribution of sounds in the text are analyzed by using the Soundex algorithm, including the measurement of change in the sounds throughout the text. The use of numbers in the text is another aspect that is measured. Additionally, words related to a pre-defined ned collection of topics were also analyzed to determine the frequency of different topics discussed in the text. All text numerical content descriptors are described in [Shamir, 2020].

Once the numerical text content descriptors are computed, machine learning is used to perform a regression based on the year of each play, as done in [Shamir, 2011]. That is done by first ranking the different descriptors by their Pearson correlation with the year in which the play was written. A descriptors that their values computed from the different plays also have strong correlation with the year are considered as descriptors that change over time. The Pearson correlation between the values and the year in which each play is written are used as weights [Shamir, 2020]. Once each text descriptor is assigned with a weight, the weighted nearest neighbor method is applied such that the Pearson correlations were used as the weights [Shamir, 2011, 2020]. That allows to predict the year of each play as reflected by the text descriptors.

## III RESULTS

---

[1] https://www.rsc.org.uk/shakespeares-plays/timeline

The method described in Section II was applied to the Shakespeare plays text files described in Section III. For each play, the predicted year was determined by the algorithm, and the results are displayed in Figure 1.

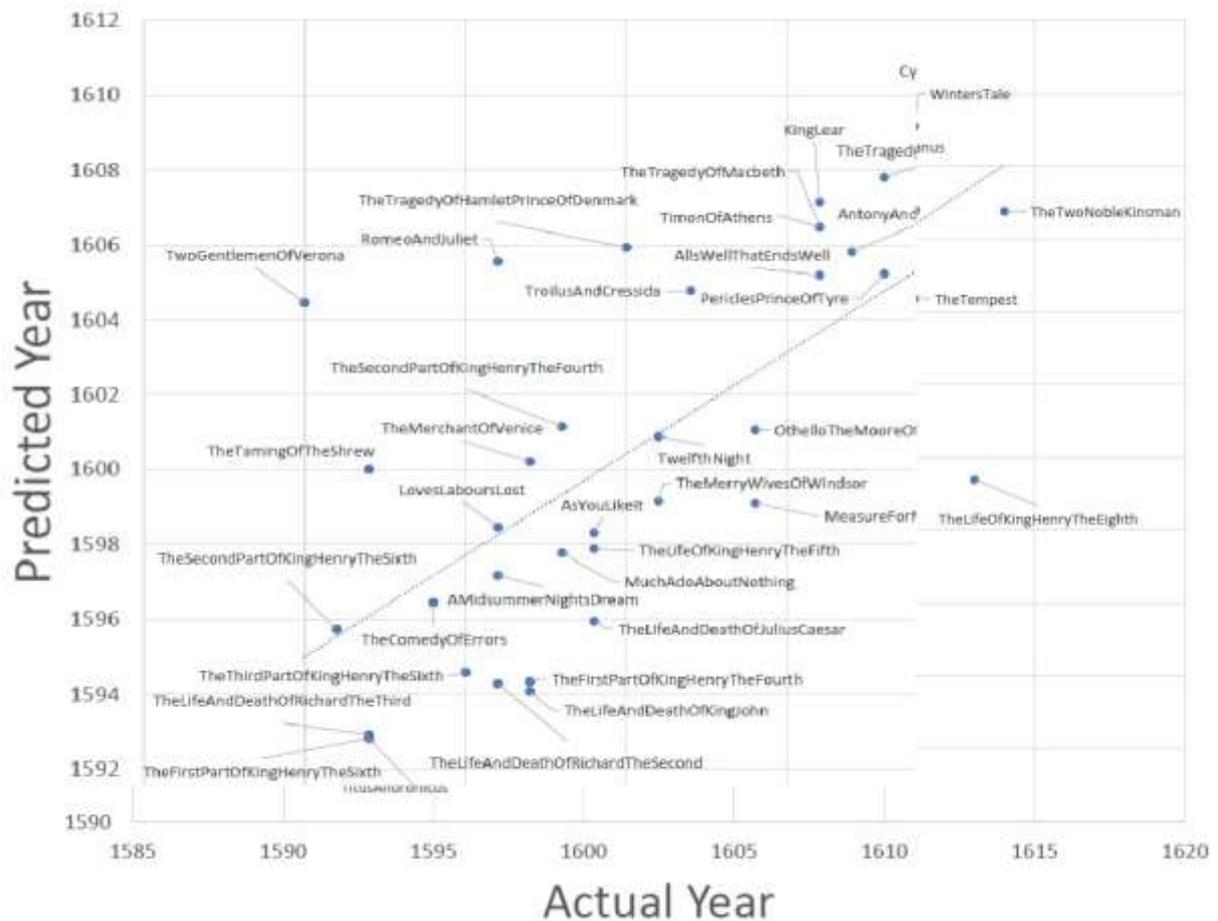

Figure 1. Estimated actual year of Shakespeare's plays (x-axis) and the predicted year of the plays (y-axis) as determined by the machine learning analysis.

As the figure shows, the predicted years of the plays as determined by the algorithm correlate with the estimated actual years the plays are assumed to be written. The Pearson correlation between the predicted year of the play and the actual year is 0.71. The two-tailed P value of the correlation is $\sim 6 \cdot 10^{-7}$. That shows strong statistical signal that indicates on changes in the style of Shakespeare over time.

To better identify specific elements that changed over time in Shakespeare's work, several different text element were examined. Figure 2 shows the change in the mean sentence length over time. As the figure shows, the sentence length mean of Shakespeare's plays generally decreased over time. The Pearson correlation coefficient between the year and the average sentence length is -0.53 (P < 0.0006). The figure also shows that sentences in Shakespeare's plays were shortest between 1600–1607, after which the sentences started to become somewhat longer.

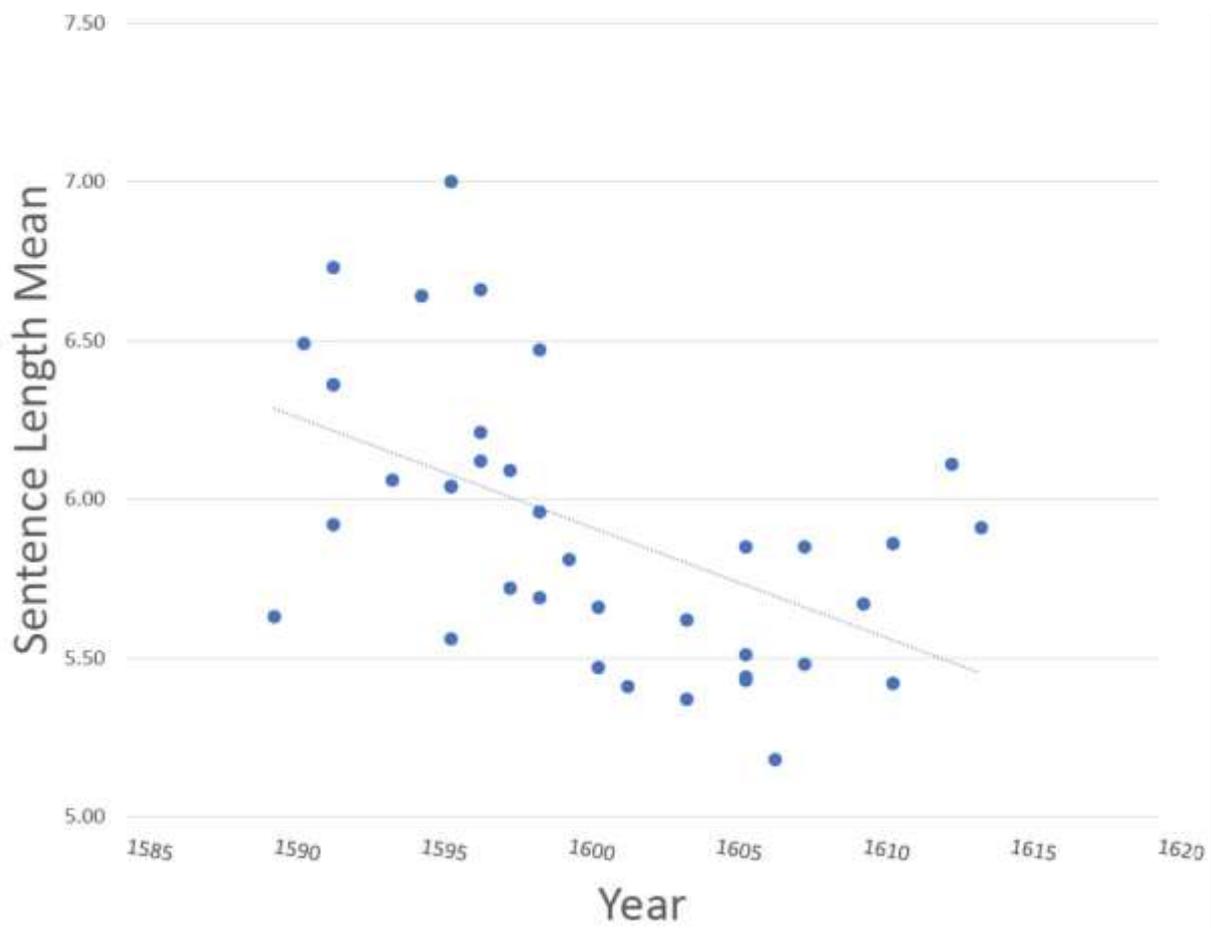

Figure 2. Sentence length mean of Shakespeare's plays in different years.

Each sentence in each play was assigned with its sentiment score as determined by the sentiment analysis used by CoreNLP [Socher et al., 2013]. The sentiment scores are between 0 and 4, with 0 being very negative, 1 being negative, 2 being neutral, 3 being positive, and 4 being very positive (Shamir, 2020). The frequency of each sentiment category was measured for each play.

The frequencies of sentences with sentiment very negative and negative over time are shown in Figure 3 and Figure 4, respectively. The frequency of sentences with sentiment very negative trends downward as shown in Figure 3, and has a Pearson correlation coefficient of -0.47 ($P < 0.003$). The frequency of sentiment negative also becomes less frequently over time, as shown by the downward trend in Figure 4 with a Pearson correlation coefficient of -0.57 ($P < 0.0002$). This indicates that Shakespeare used more negative language in his early works, and less negative language in his later works.

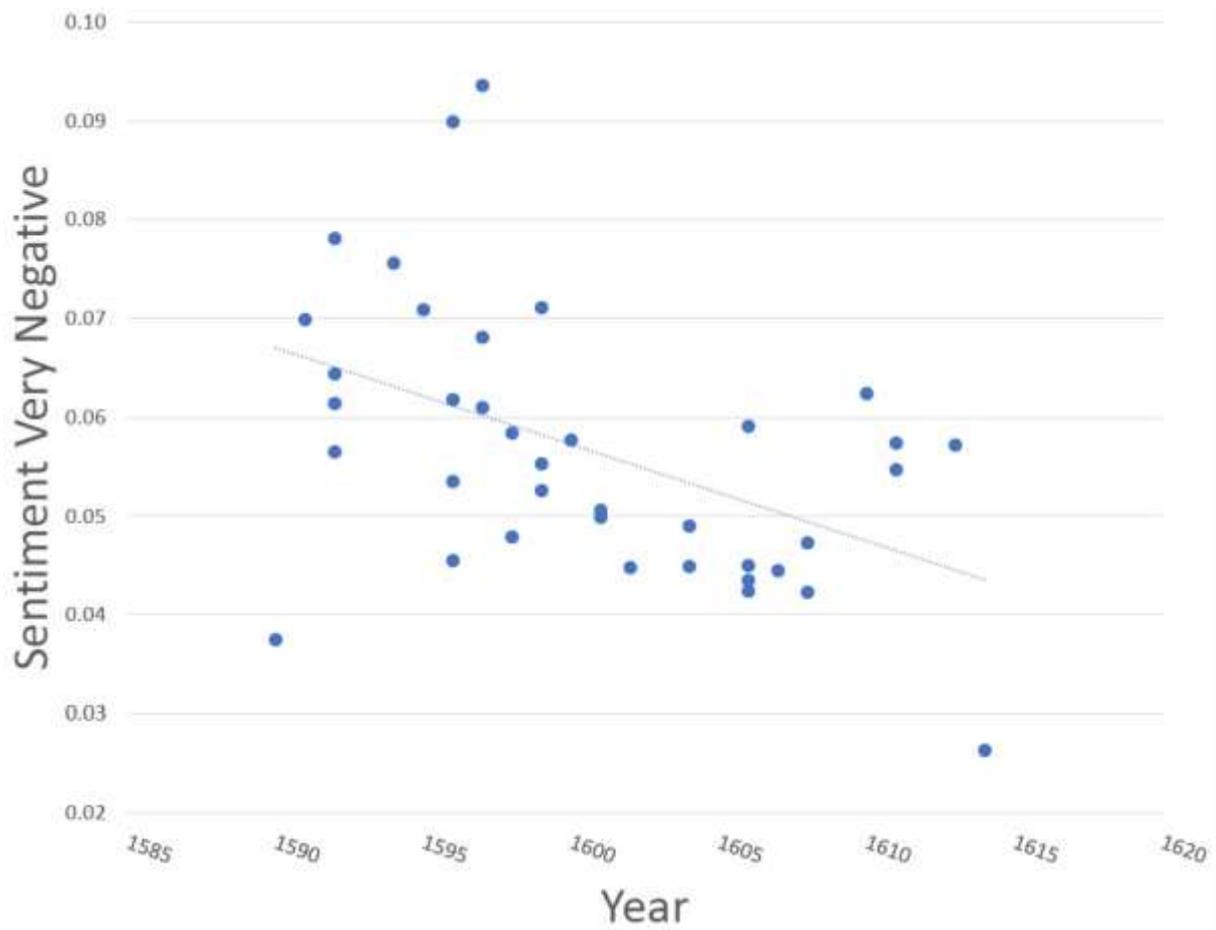

Figure 3. The frequency of sentences with sentiment *very negative* in Shakespeare's plays in different years.

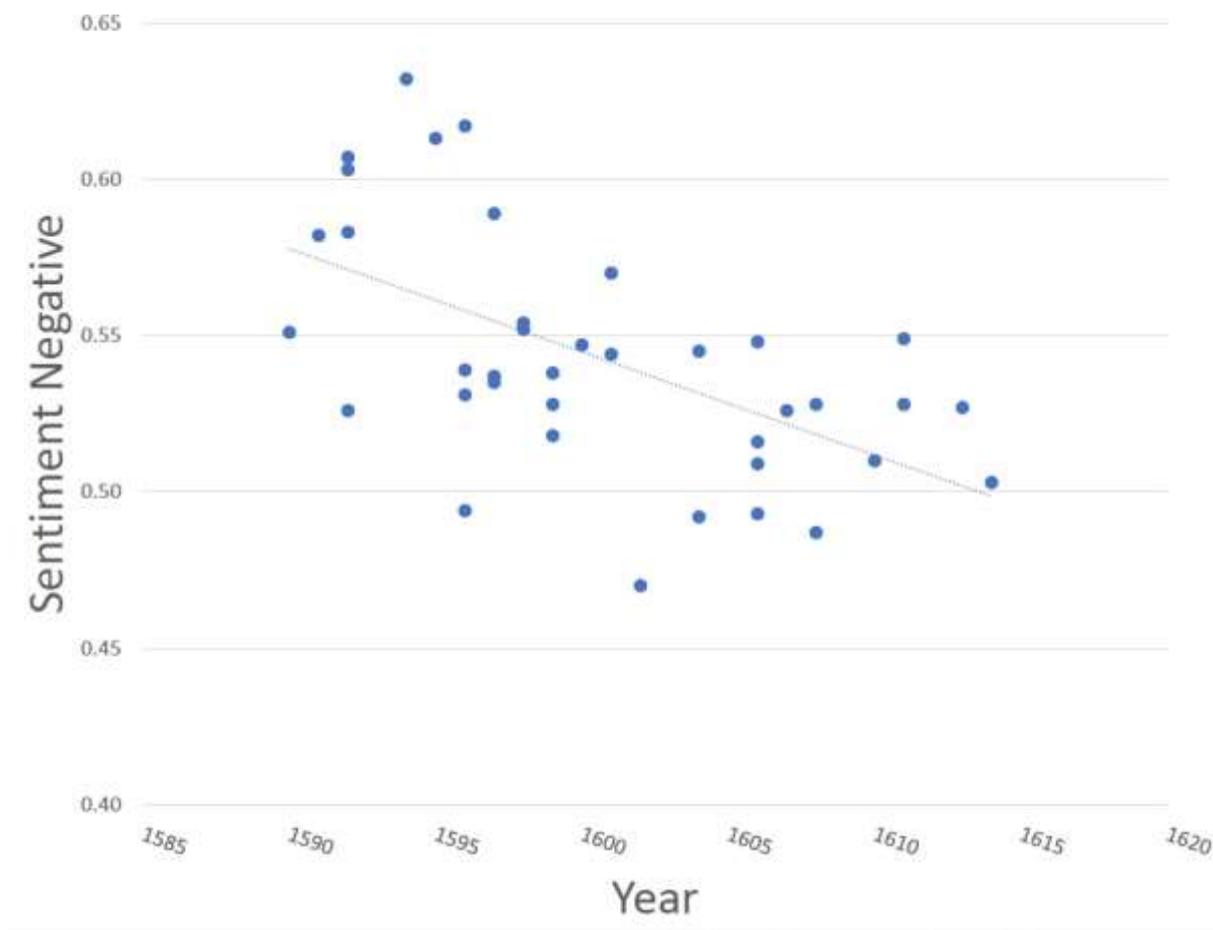

Figure 4. The frequency of sentences with sentiment *negative* in Shakespeare's plays in different years.

Figure 5 shows the frequency of sentences with positive sentiment. As the figure shows, the frequency of positive sentences increases gradually, with a Pearson correlation coefficient of 0.51 (P < 0.001). This indicates that Shakespeare's sentences expressed more positive sentiments in his later works, and his use of positive language increased over the course of his career. That observation agrees with the higher negativity in Shakespeare's earlier work as also shown in Figures 3 and 4, although the negative and positive sentiments are two independent measurements that do not directly affect each other. That is, one play can express both positive and negative sentiments, or express more negative sentiments without necessarily expressing less positive sentiments and vice versa. As the graph also shows, the trend of increasing positive sentiments over time is driven primarily by the less positive sentiments expressed in Shakespeare's early work. After around 1600 the frequency of positive sentences remains stable.

The difference can be attributed to changes in Shakespeare's late style, which was influenced through collaborations with other authors such as John Fletcher, and shifted from his previous comedy, tragedy, and history style [McDonald, 2006]. On the other hand, earlier Shakespeare plays also showed a different styles, such as the rhyming iambic pentameter used by Shakespeare in plays written in 1594–1595 [Harbage, 1962]. Stylistic and linguistic analysis such as the use of certain rare words, rhymes, and the use of colloquialism-in-verse were also used to determine the exact date of creation of Shakespeare's plays [Taylor, 1987; Whitworth, 2003], indicating that Shakespeare's style has changed over time [Klaussner and Vogel, 2015; Klaussner, 2018; Klaussner and Vogel, 2018; Andreev, 2019; Seminck et al., 2021, 2022]. While these elements are not necessarily related to sentiments and other elements tested in this study, they are in agreement with the contention that the plays of Shakespeare exhibited differences in their style in different years. A quantitative analysis showing differences between plays written before and after 1600 are provided in Section 3.1.

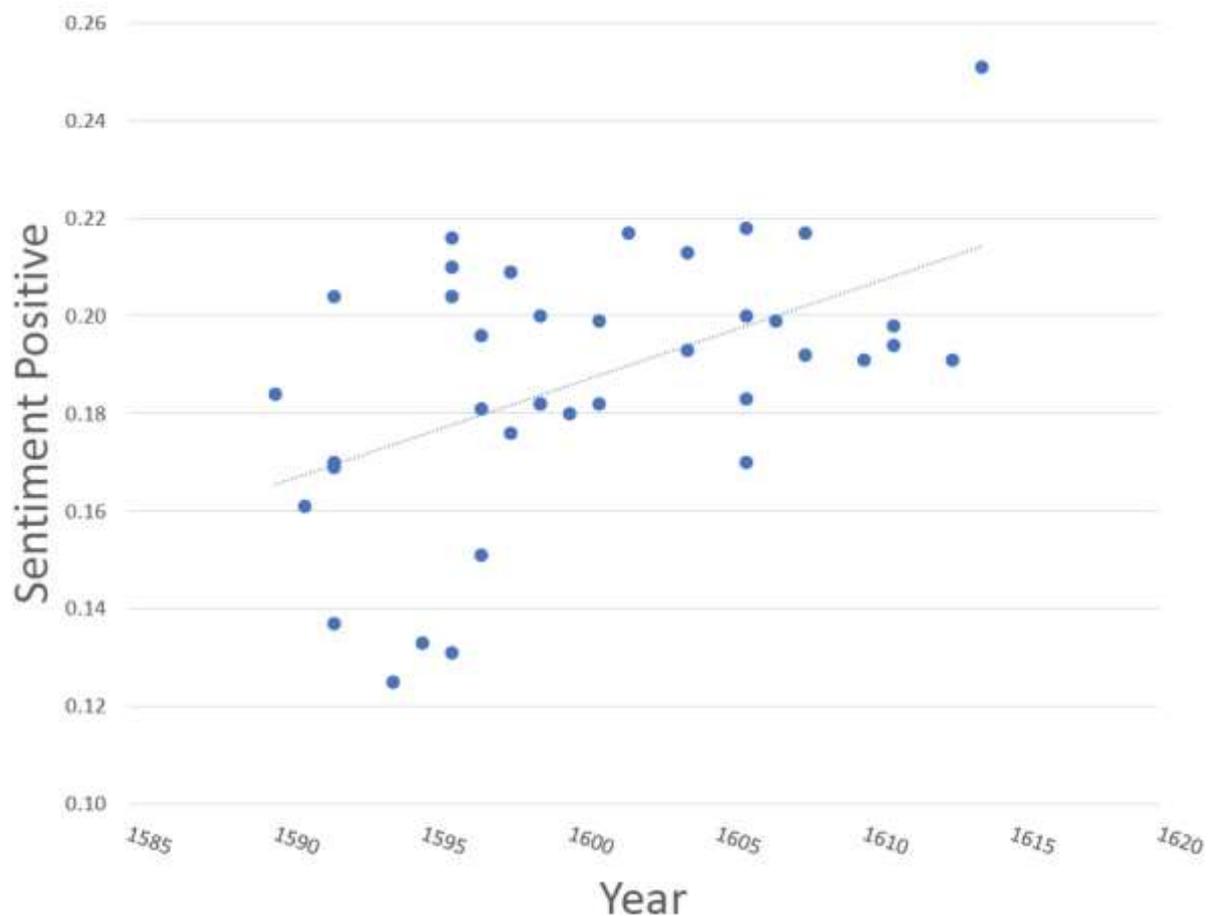

Figure 5. The frequency of sentences with sentiment *positive* in Shakespeare's plays in different years.

Lastly, Figure 6 shows the mean sentiment of the sentences in Shakespeare's plays. The Pearson correlation coefficient between the overall sentiment mean and the year is 0.58 (P < 0.0001). This shows an upward trend and, along with the figures above, shows that the type of language Shakespeare used in his early works was more negative compared to his later plays. The play with the lowest sentiment score is "The Life and Death of Richard II", written in 1596. The play with the highest sentiment score is "The Two Noble Kinsman", written in 1614.

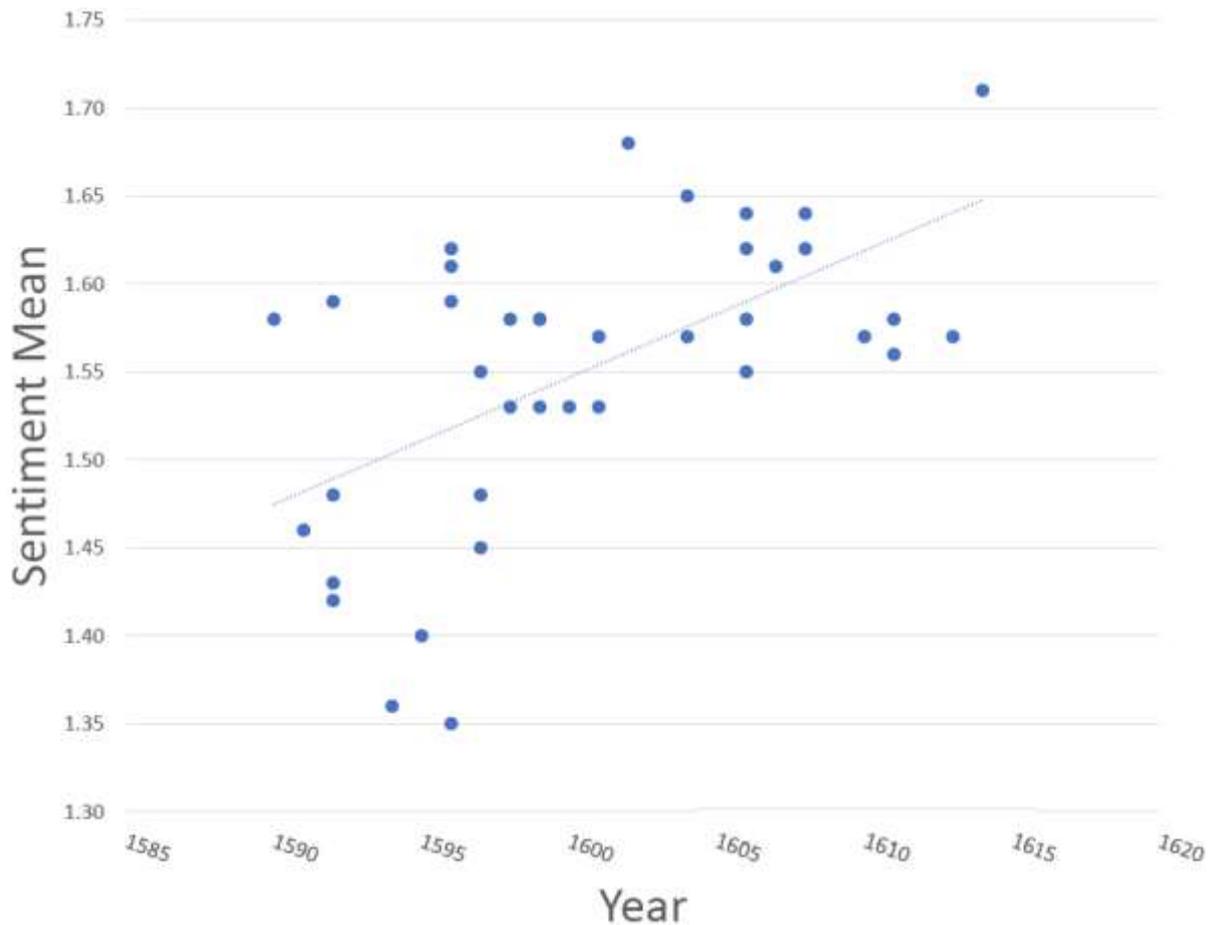

Figure 6. The sentiment mean of Shakespeare's plays in different years.

Figure 7 shows the frequency of adjective usage over time. The figure shows that Shakespeare's use of adjectives increased over time, with a Pearson cor- relation coefficient of 0.34 (P < 0.037). "The Merry Wives of Windsor" and "Twelfth Night", both written in 1601, have the lowest frequency of adjectives with scores of .055 and .061.

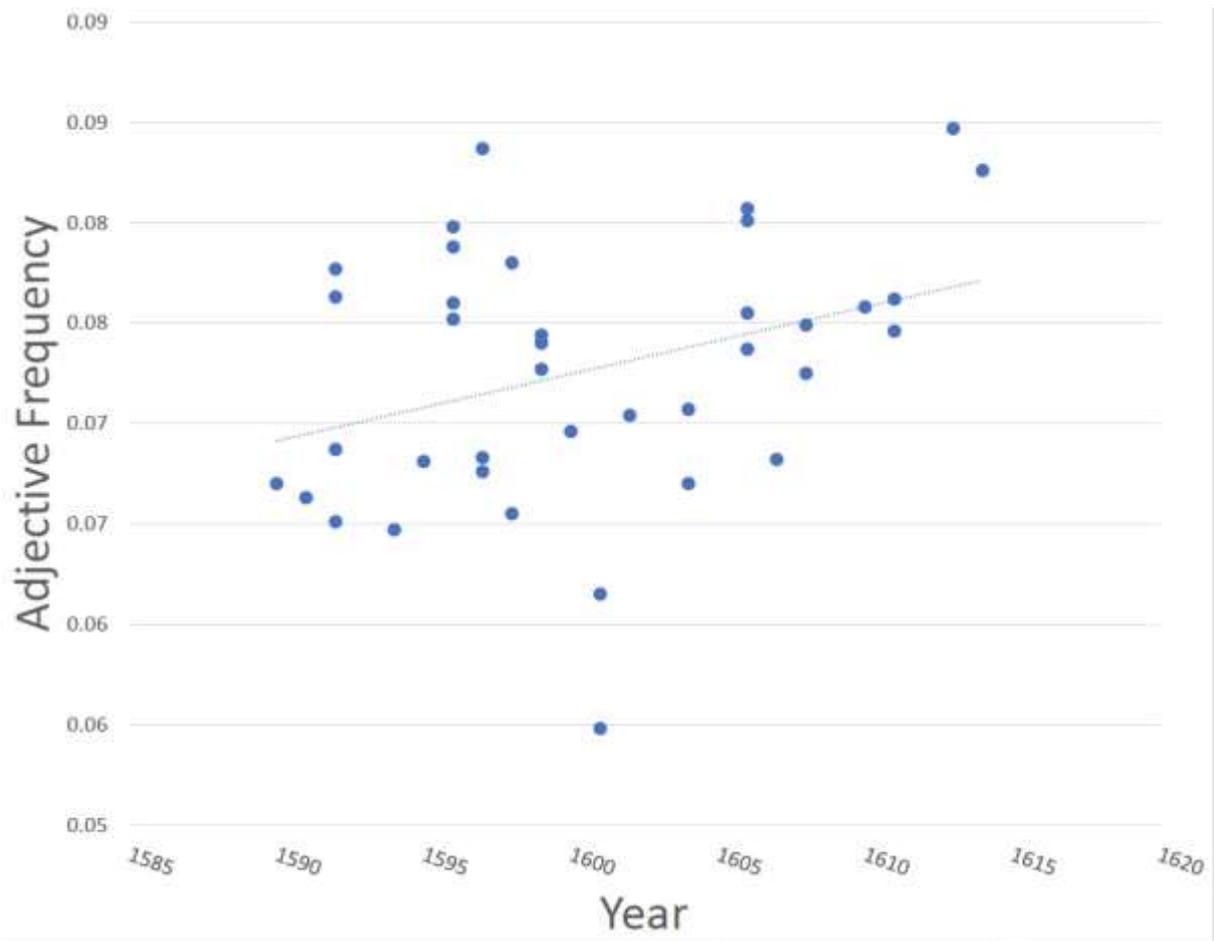

Figure 7. The adjective frequency of Shakespeare's plays in different years.

Related, the frequency of comparative and superlative adjectives increased over time, as shown by Figures 8 and 9, with Pearson correlation coefficients of 0.54 (P < 0.0005) and 0.47 (P < 0.003), respectively. The figures indicate that Shakespeare used more adjectives in his later work compared to his earlier plays.

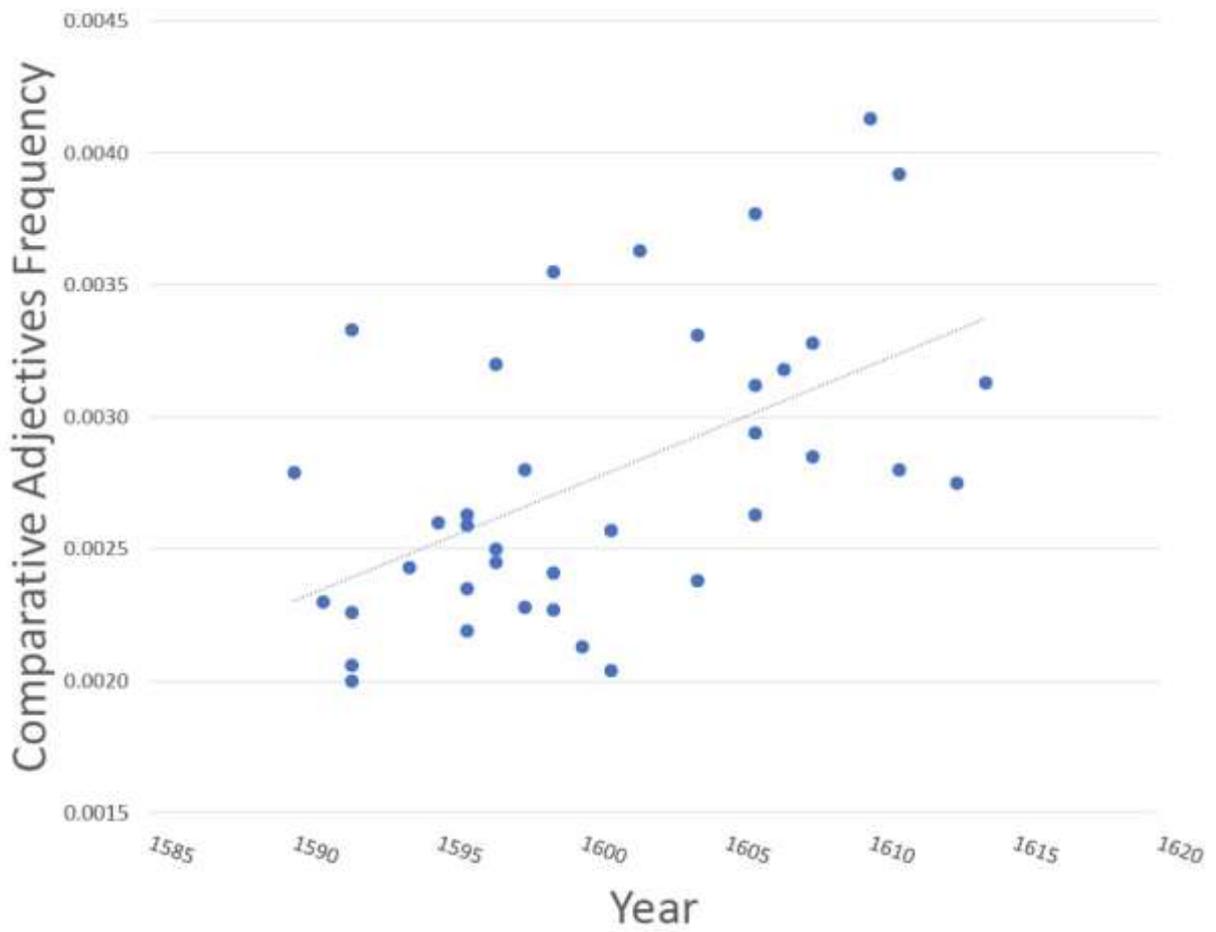

Figure 8. The comparative adjective frequency of Shakespeare's plays in different years.

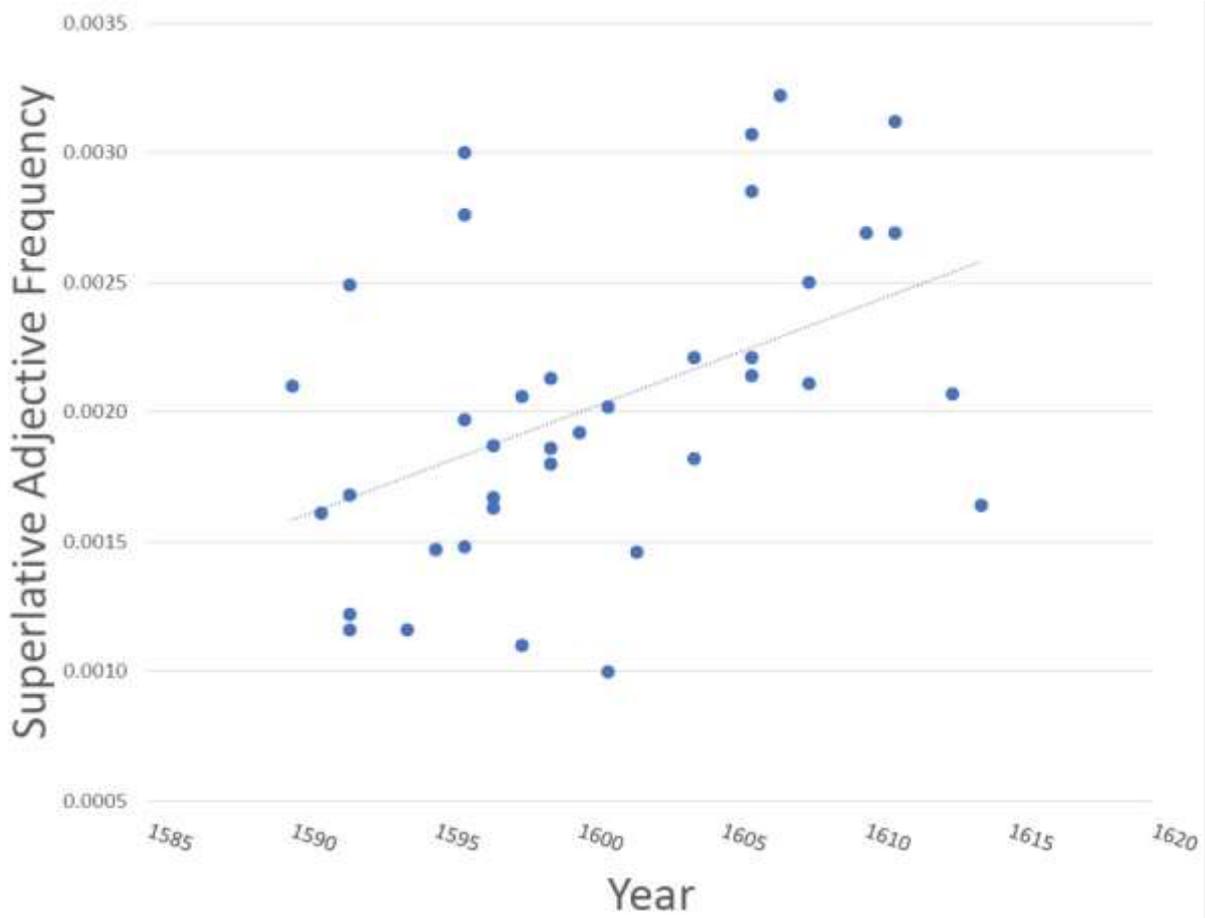

Figure 9. The superlative adjective frequency of Shakespeare's plays in different years.

The frequency of the use of adverbs also generally increased. Figure 10 and Figure 11 show the frequency of comparative and superlative adverbs over time. The Pearson correlation coefficients are 0.57 ($P < 0.0002$) and 0.66 ($P < 0.00001$), respectively. These figures suggest that more adverbs were used over the course of Shakespeare's works.

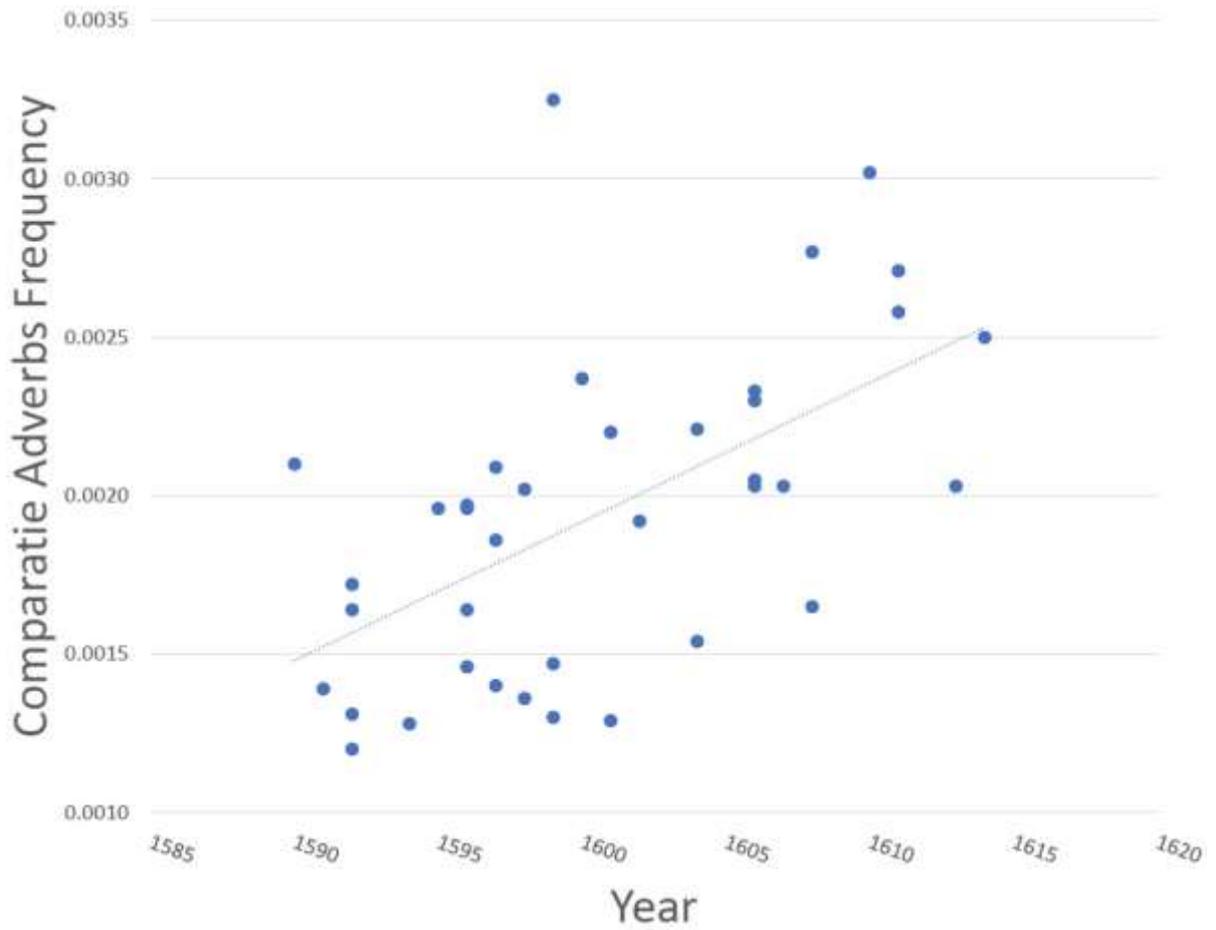
Figure 10. The comparative adverb frequency of Shakespeare's plays in different years.

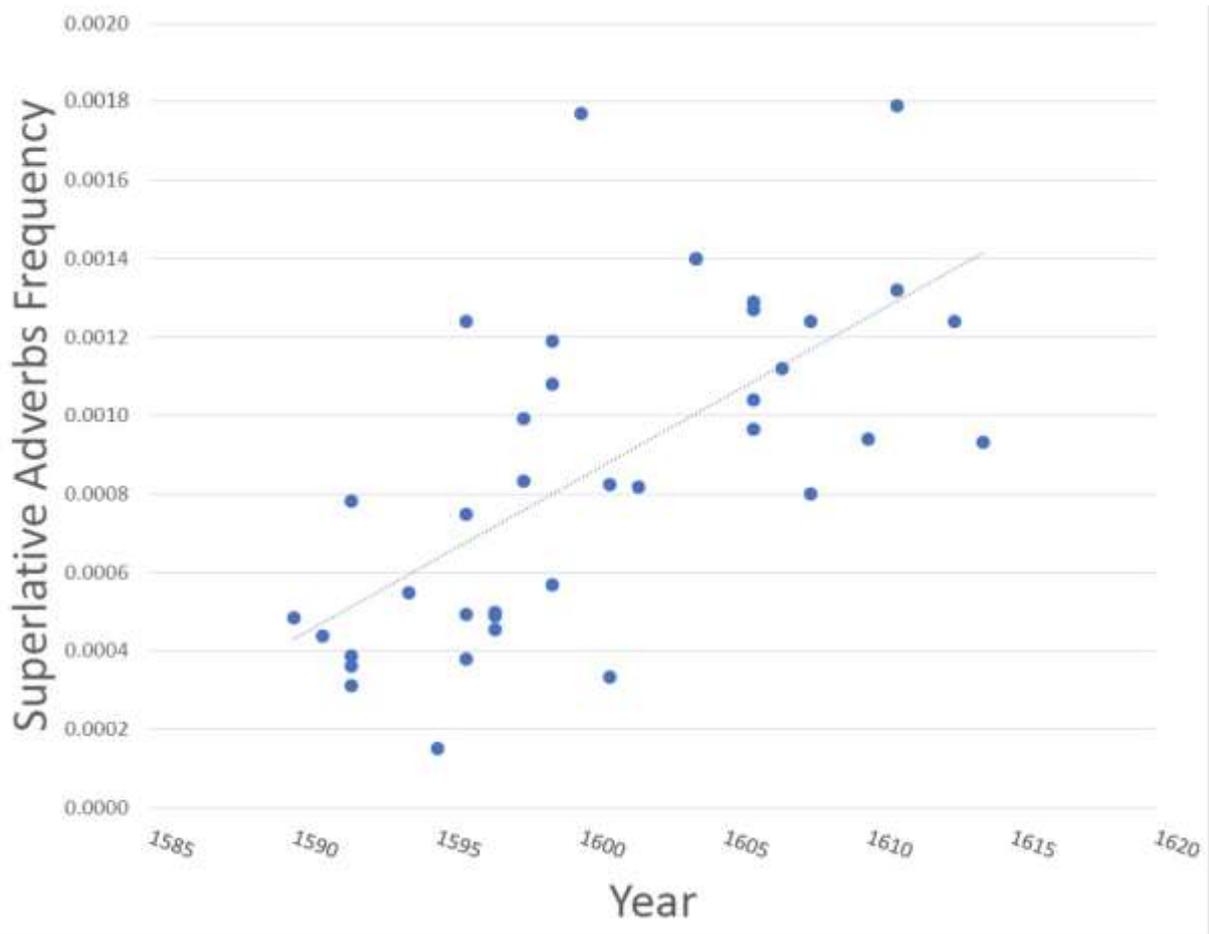

Figure 11. The superlative adverb frequency of Shakespeare's plays in different years.

Other changes include a drop in the use of conjunctions over time, as shown in Figure 12, with Pearson correlation coefficient of -0.40 (P < 0.013). The use of pronouns becomes more frequent over time as shown by Figure 13, with a Pearson correlation coefficient of 0.58 (P < 0.0001).

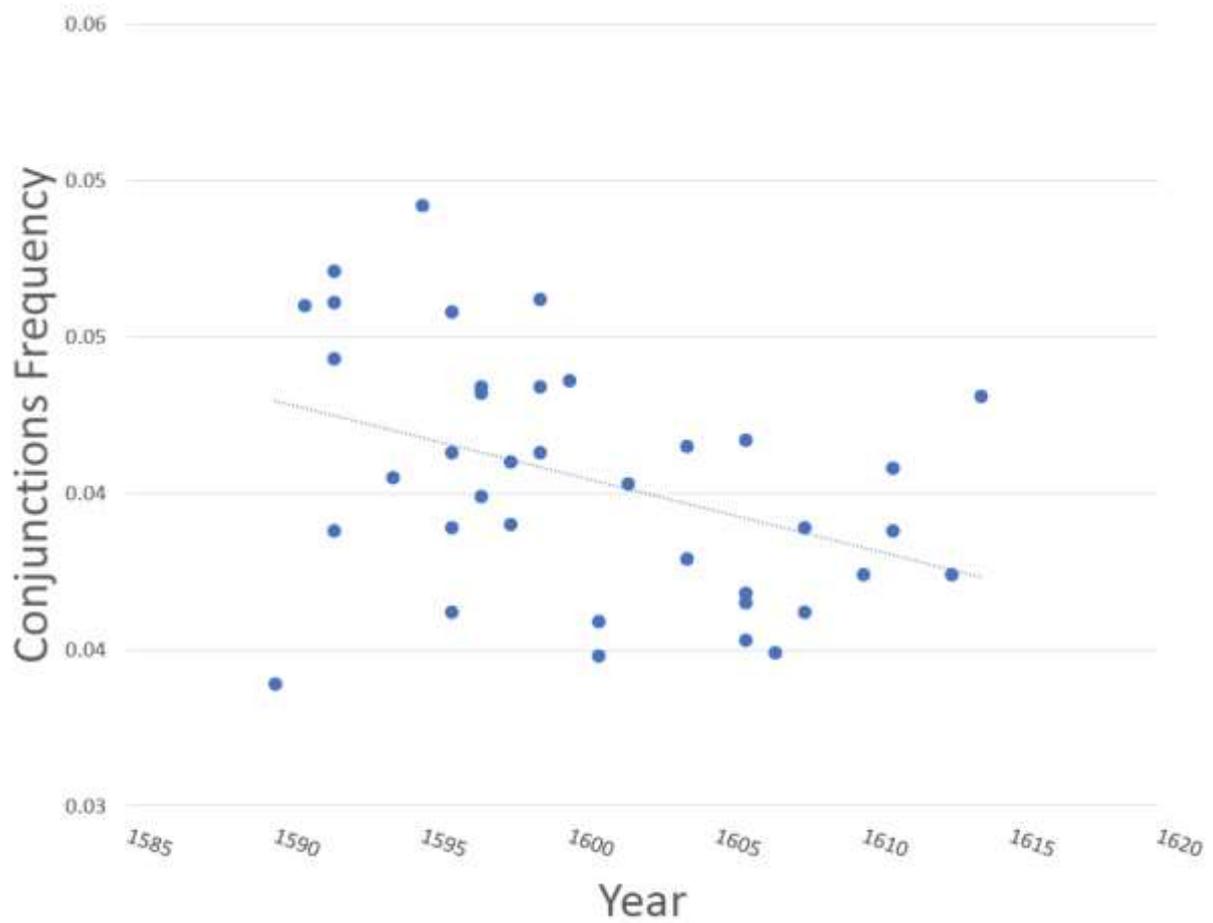

Figure 12. The conjunction frequency of Shakespeare's plays in different years.

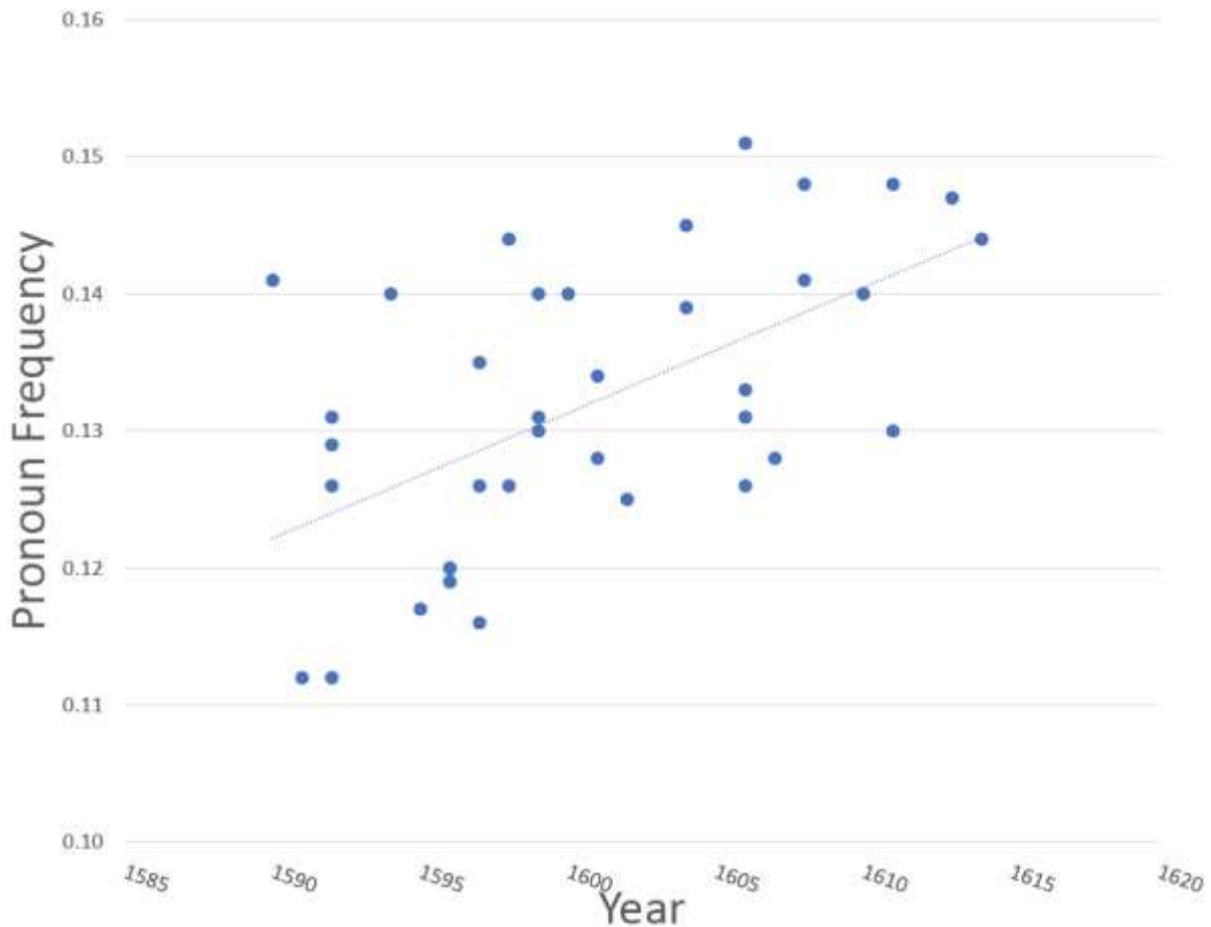

Figure 13. The pronoun frequency of Shakespeare's plays in different years.

### 3.1 Romeo and Juliet as a "precursor" play

As Figure 1 shows, some plays were predicted by the algorithm as written in a later year compared to the actual estimated year of creation. That can indicate that the style as analyzed by the computer is more typical to the writing style of Shakespeare's later work. A notable example as shown in the Figure 1 is "Romeo and Juliet", which is estimated to be written around 1596, but predicted by the algorithm as written in 1606.

To further investigate into the specific writing elements that made the computer predict 'Romeo and Juliet" as written at a later year, we compared the values of the numerical text content descriptors of "Romeo and Juliet" to the values of the early Shakespeare's plays (before 1600), and later Shakespeare's plays (after 1600). The text elements discussed can be found in Table 1.

| Feature | Mean before 1600 | Romeo and Juliet | Mean after 1600 |
|---|---|---|---|
| Word homogeneity mean | .00648±.00014 | 0.00614 | .006205±.00019 |
| Fall frequency | .000926±.000046 | 0.00116 | .00112±.00012 |
| **Summer frequency** | .0025±.000192 | 0.00299 | .00289±.00036 |
| **Weather frequency** | .0105±.000596 | 0.0132 | .0128±.00079 |
| **Modal auxiliaries frequency** | .0277±.0277 | 0.0291 | .029±.00054 |
| **Sentiment mean** | 1.497±.0211 | 1.61 | 1.607±.0140 |

Table 1. The mean of the features before and after 1600 with standard error.

The word homogeneity [Shamir, 2020] was generally higher in plays written before 1600 compared to plays written after 1600, with an average of 0.0065 before 1600 and an average of 0.0062 after 1600. The decrease suggests that Shakespeare used a more diverse selection of words and repeated the same words less often in plays written in the second half of his career. As the table shows, "Romeo and Juliet"'s word homogeneity mean is similar to plays that were written later in Shakespeare's life. The sigma difference of the word homogeneity in "Romeo and Juliet" and word homogeneity in plays written before 1600 is 2.46 sigma, while the difference from plays written after 1600 is 0.34. That shows a statistically significant difference between "Romeo and Juliet" and plays written before 1600, while the difference between "Romeo and Juliet" and plays written after 1600 is not statistically significant.

The use of words related to fall, summer, and weather are all used more frequently after 1600 compared to before 1600, with averages of 0.00093, 0.0025, 0.01 before 1600, and averages of 0.0011, 0.0029, and 0.01 after 1600. The table indicates that the usage of topic words related to fall, summer, and weather in "Romeo and Juliet" is similar to works that would be written later than compared to works written chronologically closer to the play. The sigma difference of these measurements from "Romeo and Juliet" and the plays before and after 1600 are 5.08 and 0.31, 2.45 and 0.28, and 4.50 and 0.49, respectively. That shows a statistically significant similarity between "Romeo and Juliet" and plays written after 1600.

"Romeo and Juliet" also exhibits usage of modal auxiliaries that is more similar to plays written after 1600 than before 1600. The average modal auxiliaries frequency is 0.028 before 1600, and 0.029 after 1600. The statistical strength of the difference between the frequency of modal auxiliaries in "Romeo and Juliet" and plays written before 1600 is 3.04 sigma, while the difference between "Romeo and Juliet" and plays written after 1600 is just 0.15 sigma. This shows statistically significant difference between "Romeo and Juliet" and plays written before 1600, while similarity to plays written after 1600.

Lastly, the sentiment of "Romeo and Juliet" is more positive than the plays that are written chronologically close to it. The average sentiment mean of plays written before 1600 is 1.50, and 1.61 for plays written after 1600. The sigma difference is of 5.35 and 0.20, showing that the sentiments are close to Shakespeare's later work than his earlier work. The combination of all text measurements made the machine learning algorithm predict "Romeo and Juliet" as a play written much later than its estimated year, indicating the "Romeo and Juliet" might be a precursor of Shakespeare later style.

### 3.2 Differences between comedy and tragedy
Shakespeare's work includes comedy and tragedy plays. Since the writing style of a comedy might be different from the writing style of a tragedy, comedy and tragedy might have Journal of Data Mining and Digital Humanities                                                    http://jdmdh.episciences.org
ISSN 2416-5999, an open-access journal

substantial differences between them exhibited through the style of the writing. To analyze the differences between comedy and tragedy, we used the method described in Section II and in [Shamir, 2020] to separate automatically between the comedy and tragedy plays of William Shakespeare.

The dataset included 13 comedy plays and 10 tragedies. The tragedies include Antony and Cleopatra, The tragedy of Coriolanus, Hamlet, Julius Caesar, The tragedy of King Lear, Macbeth, Othello, Romeo and Juliet, Timon of Athens, and Titus Andronicus. The comedy plays are All's Well That Ends Well, As You Like It, The Comedy of Errors, Love's Labour's Lost, Measure for Measure, The Merry Wives of Windsor, A Midsummer Night's Dream, Much Ado About Nothing, The Taming of the Shrew, The Tempest, Twelfth Night,
The Two Gentlemen of Verona, The Winter's Tale.

The first analysis that was applied was automatic classification. Nine plays from each set was used for training, and one play for testing, and the analysis was repeated 100 times such that in each run different plays were allocated for training and testing. That process is explained in detail, including the command that executes it, in [Shamir, 2020]. The results show that the algorithm was able to identify whether the play is tragedy or comedy in 74% of the cases. That accuracy is not perfect, but since it is higher than 50% it suggests that some differences exist between Shakespeare's comedy and tragedy plays.

Table 2 shows text elements that exhibit differences between comedy and tragedy plays. As the table shows, several text elements show statistically significant difference between Shakespeare's comedy plays and Shakespeare's tragedy plays. While the difference in the sentiments is expected, the table also shows that comedy plays tend to use longer words compared to tragedies. The Coleman–Liau index in comedies is also higher than in tragedies, indicating that comedies are slightly more difficult to read, and the Soundex diversity indicates that the sounds of the words used in comedies is more diverse. Shakespeare's comedies also have more questions in them, and tend to be more homogeneous in the selection of words. Another difference is the sentence length, where tragedies tend to use a more consistent length of sentences, while in comedies the length of the sentences tends to vary more.

| Feature | Mean comedy | Mean tragedy | P value |
|---|---|---|---|
| Word length mean | 3.798±.029 | 3.707±.03 | 0.042 |
| Sentiment mean | 1.761±.04 | 1.41±.04 | $<10^{-5}$ |
| Coleman-Liau index | 1.337±.18 | 0.727±.19 | 0.03 |
| Use of numbers | .0055±.0004 | .004±.0004 | 0.0162 |
| Soundex diversity | .0545±.001395 | .048±.002 | $<10^{-5}$ |
| Frequency of '?' | 0.0108±.0004 | 0.01327±.0009 | 0.025 |
| Sentence length stddev | 3.839±.092922 | 3.417±.045052 | 0.0005 |
| Word homogeneity mean | 0.00728±.0002 | 0.0065±.00016 | 0.007 |

Table 2. The mean of the text elements in comedy or tragedy. The p-values are the t-test P values of the two mean to have such difference by chance.

**Conclusion**

In the recent years, machine learning has been used for multiple tasks of text analysis to allow complex analysis of text. These include text generation such as automatic document classification [Kadhim, 2019; Kowsari et al., 2019], text generation [Gatt and Krahmer, 2018; de Rosa and Papa, 2021], text summarization [Gambhir and Gupta, 2017; El-Kassas et al., 2021], sentiment analysis [Zhang et al., 2018; Yadav and Vishwakarma, 2020], automatic caption generation [Bai and An, 2018; Hossain et al., 2019; Predić et al., 2022].

Advancements in machine learning and text analysis have been providing new ways of analyzing literature, and has been a growing sub-field in the digital humanities. Here we applied machine learning and data science techniques to analyze the changes in Shakespeare's style over time. The results show that the way Shakespeare used different grammatical structures shifted over the course of his career. For instance, sentences generally became shorter, his writing became more descriptive with a higher frequency of adjectives and adverbs, and the sentiment of the sentences in his plays became less negative.

The analysis also shows that the play "Romeo and Juliet" can be observed as a precursor of Shakespeare's stylometrics, with quantitative elements typical to Shakespeare's later work. Elements such as the word homogeneity, the usage of certain topic words, the modal auxiliaries, and sentiments is more similar to the plays written in the later half of Shakespeare's career compared to when "Romeo and Juliet" was estimated to be written.

Using machine learning analysis of Shakespeare's style is an attempt to understand the work of one of the most important authors in history. Clearly, the writing style of an author cannot be fully reduced into numerical elements, and therefore not all possible changes over the years can be captured in a mathematical analysis. Also, the analysis is not sensitive to the topics, but to the stylometrics only, and plays that share similar topics might not be identified as related to each other as long as their writing style elements are not similar.

Topics like geographic locations, occupations of the characters, and certain objects that are used, are not reflected in the analysis. Another limitation is the sentiment analysis, which is still not fully accurate, and might not always be able to capture the sentiments in a fully accurate manner. Clearly, the analysis shown here is relevant only to literature in the English language, and cannot be applied to literature in any other language.

Due to its ability to analyze a high number of dimensions, machine learning is able to identify complex patterns that are difficult to identify by manual inspection of the text. While such analysis cannot be considered complete, machine learning and statistical analyses can assist to reveal changes in the style of Shakespeare or other authors that are extremely difficult to identify without using automation. Future work can apply similar analysis to a broad variety of authors other than William Shakespeare. Instead of comparing books of the same author, the analysis can be expanded to analysis of different authors.

Since the code and software used in this work is publicly available, such work can be done without the need to have experience in programming or machine learning. The existing software does not support other languages than English, making it limited to English authors. Future work to expand the method to other languages can allow analysis of literature in other languages.